\documentclass[11pt]{article}

\usepackage{naaclhlt2016}

\usepackage{times}
\usepackage{natbib}
\usepackage{url}
\usepackage{latexsym}
\usepackage{multirow}
\usepackage{amssymb}
\usepackage{amsthm}
\usepackage{color}
\usepackage{titlesec}

\titleformat*{\subsubsection}{\LARGE\bfseries}

\usepackage{amsmath}
\usepackage{verbatim}
\usepackage{tikz-qtree}
\usepackage{tikz}
\usetikzlibrary{shapes.geometric}
\usetikzlibrary{shapes.arrows}
\usetikzlibrary{patterns}
\usetikzlibrary{chains}
\usetikzlibrary{calc}

\usepackage{framed}

\newenvironment{itemizesquish}{\begin{list}{\setcounter{enumi}{0}\labelitemi}{\setlength{\itemsep}{-0.25em}\setlength{\labelwidth}{0.5em}\setlength{\leftmargin}{\labelwidth}\addtolength{\leftmargin}{\labelsep}}}{\end{list}}

\newcommand{\ignore}[1]{}

\newcommand{\ssep}{\,\mid\,}

\naaclfinalcopy

\newsavebox{\one}
\newsavebox{\two}
\newsavebox{\three}
\newsavebox{\four}
\newsavebox{\five}

\title{Recurrent Neural Network Grammars}
\author{Chris Dyer$^{\spadesuit}$ ~ Adhiguna Kuncoro$^{\spadesuit}$ ~ Miguel Ballesteros$^{\diamondsuit\spadesuit}$ ~ Noah A. Smith$^{\heartsuit}$\\
$^{\spadesuit}$School of Computer Science, Carnegie Mellon University, Pittsburgh, PA, USA \\
$^{\diamondsuit}$NLP Group, Pompeu Fabra University, Barcelona, Spain \\
$^{\heartsuit}$Computer Science \& Engineering, University of Washington, Seattle, WA, USA\\
{\small \tt \{cdyer,akuncoro\}@cs.cmu.edu, miguel.ballesteros@upf.edu, nasmith@cs.washington.edu}
}
\date{}

\begin{document}

\maketitle

\begin{framed}
\noindent This is modified version of a paper originally published at NAACL 2016 that contains a corrigendum at the end, with improved results after fixing an implementation bug in the RNNG composition function.
\end{framed}

\begin{abstract}
 We introduce recurrent neural network grammars, probabilistic models of sentences with explicit phrase structure.  We explain efficient inference procedures that allow application to
both parsing and language modeling.  Experiments show that they
provide better parsing in English than any single previously published supervised generative model and
better language modeling than state-of-the-art sequential RNNs in English and Chinese\footnote{The code to reproduce our results after the bug fix is publicly available at \url{https://github.com/clab/rnng}.}.
\end{abstract}

\section{Introduction}
Sequential recurrent neural networks (RNNs) are remarkably effective models of natural language. In the last few years, language model results that substantially improve over long-established state-of-the-art baselines have been obtained using RNNs \citep{zaremba:2015b,mikolov:2010} as well as in various conditional language modeling tasks such as machine translation \citep{bahdanau:2015}, image caption generation \citep{xu:2015}, and dialogue generation \citep{wen:2015}. Despite these impressive results, sequential models are \emph{a priori} inappropriate models of natural language, since relationships among words are largely organized in terms of latent nested structures rather than sequential surface order~\citep{chomsky:1957}.

In this paper, we introduce \textbf{recurrent neural network grammars} (RNNGs; \S\ref{sec:formal}), a new generative probabilistic model of sentences that explicitly models nested, hierarchical relationships among words and phrases. RNNGs operate via a recursive syntactic process reminiscent of probabilistic context-free grammar generation, but decisions are parameterized using RNNs that condition on the entire syntactic derivation history, greatly relaxing context-free independence assumptions.

The foundation of this work is a top-down variant of transition-based parsing (\S\ref{sec:top-down}). We give two variants of the algorithm, one for parsing (given an observed sentence, transform it into a tree), and one for generation. While several transition-based neural models of syntactic generation exist \citep{henderson:2003,henderson:2004,emami:2005,titov:2007b,buys:2015b}, these have relied on structure building operations based on parsing actions in shift-reduce and left-corner parsers which operate in a largely bottom-up fashion. While this construction is appealing because inference is relatively straightforward, it limits the use of top-down grammar information, which is helpful for generation~\citep{roark:2001}.\footnote{The left-corner parsers used by Henderson~(2003, 2004) incorporate limited top-down information, but a complete path from the root of the tree to a terminal is not generally present when a terminal is generated. Refer to \citet[Fig. 1]{henderson:2003} for an example.} RNNGs maintain the algorithmic convenience of transition-based parsing but incorporate top-down (i.e., root-to-terminal) syntactic information~(\S\ref{sec:model}).

The top-down transition set that RNNGs are based on lends itself to discriminative modeling as well, where sequences of transitions are modeled conditional on the full input sentence along with the incrementally constructed syntactic structures. Similar to previously published discriminative bottom-up transition-based parsers~\citep[\emph{inter alia}]{henderson:2004,sagae:2005,zhang:2011}, greedy prediction with our model yields a linear-time deterministic parser (provided an upper bound on the number of actions taken between processing subsequent terminal symbols is imposed); however, our algorithm generates arbitrary tree structures directly, without the binarization required by shift-reduce parsers. The discriminative model also lets us use ancestor sampling to obtain samples of parse trees for sentences, and this is used to solve a second practical challenge with RNNGs: approximating the marginal likelihood and MAP tree of a sentence under the generative model. We present a simple importance sampling algorithm which uses samples from the discriminative parser to solve inference problems in the generative model~(\S\ref{sec:inference}).

Experiments show that RNNGs are effective for both language modeling and parsing~(\S\ref{sec:exp}). Our generative model obtains (i)~the best-known parsing results using a single supervised generative model and
(ii)~better perplexities in language modeling than state-of-the-art sequential LSTM language models. Surprisingly---although in line with previous parsing results showing the effectiveness of generative models~\citep{henderson:2004,johnson:2001}---parsing with the generative model obtains significantly better results than parsing with the discriminative model.

\section{RNN Grammars}\label{sec:formal}
Formally, an RNNG is a triple $(N, \Sigma, \Theta)$ consisting of a finite set of nonterminal symbols ($N$), a finite set of terminal symbols ($\Sigma$) such that $N \cap \Sigma = \emptyset$, and a collection of neural network parameters $\Theta$.  It does not explicitly define rules since these are implicitly characterized by $\Theta$. The algorithm that the grammar uses to generate trees and strings in the language is characterized in terms of a transition-based algorithm, which is outlined in the next section. In the section after that, the semantics of the parameters that are used to turn this into a stochastic algorithm that generates pairs of trees and strings are discussed.

\section{Top-down Parsing and Generation}\label{sec:top-down}
RNNGs are based on a top-down generation algorithm that relies on a stack data structure of partially completed syntactic constituents. To emphasize the similarity of our algorithm to more familiar bottom-up shift-reduce recognition algorithms, we first present the parsing (rather than generation) version of our algorithm~(\S\ref{sec:parse}) and then present modifications to turn it into a generator~(\S\ref{sec:gen}).

\subsection{Parser Transitions}\label{sec:parse}
The parsing algorithm transforms a sequence of words $\boldsymbol{x}$ into a parse tree $\boldsymbol{y}$ using two data structures (a stack and an input buffer). As with the bottom-up algorithm of \citet{sagae:2005}, our algorithm begins with the stack ($S$) empty and the complete sequence of words in the input buffer ($B$). The buffer contains unprocessed terminal symbols, and the stack contains terminal symbols, ``open'' nonterminal symbols, and completed constituents. At each timestep, one of the following three classes of operations (Fig.~\ref{fig:parser}) is selected by a classifier, based on the current contents on the stack and buffer:
\begin{itemizesquish}
\item $\textsc{nt}(\mathrm{X})$ introduces an ``open nonterminal'' X onto the top of the stack. Open nonterminals are written as a nonterminal symbol preceded by an open parenthesis, e.g., ``(VP'', and they represent a nonterminal whose child nodes have not yet been fully constructed. Open nonterminals are ``closed'' to form complete constituents by subsequent \textsc{reduce} operations.
\item \textsc{shift} removes the terminal symbol $x$ from the front of the input buffer, and pushes it onto the top of the stack.
\item \textsc{reduce} repeatedly pops completed subtrees or terminal symbols from the stack until an open nonterminal is encountered, and then this open NT is popped and used as the label of a new constituent that has the popped subtrees as its children. This new completed constituent is pushed onto the stack as a single composite item. A single \textsc{reduce} operation can thus create constituents with an unbounded number of children.
\end{itemizesquish}
The parsing algorithm terminates when there is a single completed constituent on the stack and the buffer is empty. Fig.~\ref{fig:example} shows an example parse using our transition set. Note that in this paper we do not model preterminal symbols (i.e., part-of-speech tags) and our examples therefore do not include them.\footnote{Preterminal symbols are, from the parsing algorithm's point of view, just another kind of nonterminal symbol that requires no special handling. However, leaving them out reduces the number of transitions by $O(n)$ and also reduces the number of action types, both of which reduce the runtime. Furthermore, standard parsing evaluation scores do not depend on preterminal prediction accuracy.}

Our transition set is closely related to the operations used in Earley's algorithm which likewise introduces nonterminals symbols with its \textsc{predict} operation and later \textsc{complete}s them after consuming terminal symbols one at a time using \textsc{scan} \citep{earley:1970}.\ignore{\footnote{The close relationship between transition sets used in incremental parsers and dynamic programming parsers has been noted in the past in the context of bottom-up parsing~\citep{huang:2010}, where it has been used to construct structured search algorithms.}} It is likewise closely related to the ``linearized'' parse trees proposed by \citet{vinyals:2015} and to the top-down, left-to-right decompositions of trees used in previous generative parsing and language modeling work~\citep{roark:2001,roark:2004,charniak:2010}.

A further connection is to $LL(^*)$ parsing which uses an unbounded lookahead (compactly represented by a DFA) to distinguish between parse alternatives in a top-down parser~\citep{parr:2011}; however, our parser uses an RNN encoding of the lookahead rather than a DFA.

\begin{figure*}
\centering
\begin{tabular}{lll|l|lll}
\textbf{Stack}$_t$ & \textbf{Buffer}$_t$ & \textbf{Open NTs}$_t$ & \textbf{Action} & \textbf{Stack}$_{t+1}$ & \textbf{Buffer}$_{t+1}$ & \textbf{Open NTs}$_{t+1}$ \\
\hline
$S$ & $B$ & $n$ &$\textsc{nt}(\mathrm{X})$ & $S\mid(\mathrm{X}$ & $B$ & $n+1$ \\
$S$ & $x\mid B$ & $n$ & $\textsc{shift}$ & $S\mid \ x$ & $B$& $n$ \\
$S\mid(\mathrm{X}\mid \tau_1\mid \ldots\mid \tau_{\ell}$ & $B$ & $n$ & \textsc{reduce} & $S\mid (\mathrm{X}\ \tau_1\ \ldots\  \tau_{\ell})$ & $B$ & $n-1$
\end{tabular}
\hspace{+.5mm}
\caption{\label{fig:parser}Parser transitions showing the stack, buffer, and open nonterminal count before and after each action type. $S$ represents the stack, which contains open nonterminals and completed subtrees; $B$ represents the buffer of unprocessed terminal symbols; $x$ is a terminal symbol, $\mathrm{X}$ is a nonterminal symbol, and each $\tau$ is a completed subtree. The top of the stack is to the right, and the buffer is consumed from left to right.  Elements on the stack and buffer are delimited by a vertical bar ($\ssep$).}
\end{figure*}

\begin{figure*}
${}$\qquad\qquad\quad\textbf{Input: }{\it The hungry cat meows .}
\begin{center}
\begin{tabular}{r|l|l|l}
&\textbf{Stack}&\textbf{Buffer}&\textbf{Action}\\
\hline
\tiny 0 & &{\it The}$\ssep${\it hungry}$\ssep${\it cat}$\ssep${\it meows}$\ssep${\it .}& \textsc{nt}(S)\\
\tiny 1  & (S&{\it The}$\ssep${\it hungry}$\ssep${\it cat}$\ssep${\it meows}$\ssep${\it .}& \textsc{nt}(NP) \\
\tiny 2  &  (S$\ssep$(NP&{\it The}$\ssep${\it hungry}$\ssep${\it cat}$\ssep${\it meows}$\ssep${\it .}& \textsc{shift} \\
\tiny 3  &  (S$\ssep$(NP$\ssep${\it The}&{\it hungry}$\ssep${\it cat}$\ssep${\it meows}$\ssep${\it .}& \textsc{shift} \\
\tiny 4  &   (S$\ssep$(NP$\ssep${\it The}$\ssep${\it hungry}&{\it cat}$\ssep${\it meows}$\ssep${\it .}& \textsc{shift} \\
\tiny 5  &  (S$\ssep$(NP$\ssep${\it The}$\ssep${\it hungry}$\ssep${\it cat}&{\it meows}$\ssep${\it .}& \textsc{reduce} \\
\tiny 6  &   (S$\ssep$(NP {\it The hungry cat})&{\it meows}$\ssep${\it .}& \textsc{nt}(VP) \\
\tiny 7  &   (S$\ssep$(NP {\it The hungry cat})$\ssep$(VP&{\it meows}$\ssep${\it .}& \textsc{shift} \\
\tiny 8  &  (S$\ssep$(NP {\it The hungry cat})$\ssep$(VP {\it meows}&{\it .} & \textsc{reduce} \\
\tiny 9  &  (S$\ssep$(NP {\it The hungry cat})$\ssep$(VP {\it meows})&{\it .} & \textsc{shift} \\
\tiny 10  &  (S$\ssep$(NP {\it The hungry cat})$\ssep$(VP {\it meows})$\ssep${\it .}& & \textsc{reduce} \\
\tiny 11  &  (S (NP {\it The hungry cat}) (VP {\it meows}) {\it .})& & \\
\hline
\end{tabular}
\end{center}
\caption{\label{fig:example}Top-down parsing example.}
\end{figure*}

\begin{figure*}
\centering
\begin{tabular}{lll|l|lll}
\textbf{Stack}$_t$ & \textbf{Terms}$_t$ & \textbf{Open NTs}$_t$ & \textbf{Action} & \textbf{Stack}$_{t+1}$ & \textbf{Terms}$_{t+1}$ & \textbf{Open NTs}$_{t+1}$ \\
\hline
$S$  & $T$ & $n$ &$\textsc{nt}(\mathrm{X})$ & $S\mid(\mathrm{X}$ & $T$ & $n+1$ \\
$S$  & $T$ & $n$ & $\textsc{gen}(x)$ & $S\mid \ x$  & $T\mid x$ & $n$ \\
$S\mid(\mathrm{X}\mid \tau_1\mid \ldots\mid \tau_{\ell}$ & $T$ & $n$ & \textsc{reduce} & $S\mid (\mathrm{X}\ \tau_1\ \ldots\  \tau_{\ell})$ & $T$ & $n-1$
\end{tabular}
\hspace{+.5mm}
\caption{\label{fig:generator}Generator transitions. Symbols defined as in Fig.~\ref{fig:parser} with the addition of $T$ representing the history of generated terminals.}
\end{figure*}

\begin{figure*}
\begin{center}
\begin{tabular}{r|l|l|l}
&\textbf{Stack}&\textbf{Terminals}&\textbf{Action}\\
\hline
\tiny 0  & && \textsc{nt}(S)\\
\tiny 1  &  (S&& \textsc{nt}(NP) \\
\tiny 2  &  (S$\ssep$(NP&& \textsc{gen}$(\textit{The})$ \\
\tiny 3  &  (S$\ssep$(NP$\ssep${\it The}&{\it The}& \textsc{gen}$(\textit{hungry})$ \\
\tiny 4  &  (S$\ssep$(NP$\ssep${\it The}$\ssep${\it hungry}& {\it The}$\ssep${\it hungry}& \textsc{gen}$(\textit{cat})$ \\
\tiny 5  &  (S$\ssep$(NP$\ssep${\it The}$\ssep${\it hungry}$\ssep${\it cat}& {\it The}$\ssep${\it hungry}$\ssep${\it cat} & \textsc{reduce} \\
\tiny 6  &  (S$\ssep$(NP {\it The hungry cat})& {\it The}$\ssep${\it hungry}$\ssep${\it cat} & \textsc{nt}(VP) \\
\tiny 7  &  (S$\ssep$(NP {\it The hungry cat})$\ssep$(VP& {\it The}$\ssep${\it hungry}$\ssep${\it cat} & \textsc{gen}$(\textit{meows})$ \\
\tiny 8  &  (S$\ssep$(NP {\it The hungry cat})$\ssep$(VP {\it meows}&{\it The}$\ssep${\it hungry}$\ssep${\it cat}$\ssep${\it meows}&\textsc{reduce} \\
\tiny 9  &  (S$\ssep$(NP {\it The hungry cat})$\ssep$(VP {\it meows})&{\it The}$\ssep${\it hungry}$\ssep${\it cat}$\ssep${\it meows}& \textsc{gen}$(\textit{.})$ \\
\tiny 10  &  (S$\ssep$(NP {\it The hungry cat})$\ssep$(VP {\it meows})$\ssep${\it .}&{\it The}$\ssep${\it hungry}$\ssep${\it cat}$\ssep${\it meows}$\ssep${\it .} &\textsc{reduce} \\
\tiny 11  &   (S (NP {\it The hungry cat}) (VP {\it meows}) {\it .})&{\it The}$\ssep${\it hungry}$\ssep${\it cat}$\ssep${\it meows}$\ssep${\it .}& \\
\hline
\end{tabular}
\end{center}
\caption{\label{fig:gen}Joint generation of a parse tree and sentence.}
\end{figure*}

\paragraph{Constraints on parser transitions.}
To guarantee that only well-formed phrase-structure trees are produced by the parser, we impose the following constraints on the transitions that can be applied at each step which are a function of the parser state $(B,S,n)$ where $n$ is the number of open nonterminals on the stack:
\begin{itemizesquish}
\item The $\textsc{nt}(\mathrm{X})$ operation can only be applied if $B$ is not empty and $n<100$.\footnote{Since our parser allows unary nonterminal productions, there are an infinite number of valid trees for finite-length sentences. The $n <100$ constraint prevents the classifier from misbehaving and generating excessively large numbers of nonterminals. Similar constraints have been proposed to deal with the analogous problem in bottom-up shift-reduce parsers \citep{sagae:2005}.}
\item The \textsc{shift} operation can only be applied if $B$ is not empty and $n \ge 1$.
\item The \textsc{reduce} operation can only be applied if the top of the stack is not an open nonterminal symbol.
\item The \textsc{reduce} operation can only be applied if $n \ge 2$ or if the buffer is empty.
\end{itemizesquish}
To designate the set of valid parser transitions, we write $\mathcal{A}_D(B,S,n)$.

\subsection{Generator Transitions}\label{sec:gen}
The parsing algorithm that maps from sequences of words to parse trees can be adapted with minor changes to produce an algorithm that stochastically generates trees and terminal symbols.  Two  changes are required: (i)~there is no input buffer of unprocessed words, rather there is an output buffer ($T$),  and (ii)~instead of a \textsc{shift} operation there are $\textsc{gen}(x)$ operations which generate  terminal symbol $x \in \Sigma$ and add it to the top of the stack and the output buffer. At each timestep an action is stochastically selected according to a conditional distribution that depends on the current contents of $S$ and $T$. The algorithm terminates when a single completed constituent remains on the stack. Fig.~\ref{fig:gen} shows an example generation sequence.
\paragraph{Constraints on generator transitions.}
The generation algorithm also requires slightly modified constraints. These are:
\begin{itemizesquish}
\item The $\textsc{gen}(x)$ operation can only be applied if $n \ge 1$.
\item The \textsc{reduce} operation can only be applied if the top of the stack is not an open nonterminal symbol and $n \ge 1$.
\end{itemizesquish}
To designate the set of valid generator transitions, we write $\mathcal{A}_G(T,S,n)$.

This transition set generates trees using nearly the same structure building actions and stack configurations as the ``top-down PDA'' construction proposed by \citet{abney:1999}, albeit without the restriction that the trees be in Chomsky normal form.

\subsection{Transition Sequences from Trees}
Any parse tree can be converted to a sequence of transitions via a depth-first, left-to-right traversal of a parse tree. Since there is a unique depth-first, left-ro-right traversal of a tree, there is exactly one transition sequence of each tree. For a tree $\boldsymbol{y}$ and a sequence of symbols $\boldsymbol{x}$, we write $\boldsymbol{a}(\boldsymbol{x},\boldsymbol{y})$ to indicate the corresponding sequence of generation transitions, and $\boldsymbol{b}(\boldsymbol{x},\boldsymbol{y})$ to indicate the parser transitions.

\subsection{Runtime Analysis}
A detailed analysis of the algorithmic properties of our top-down parser is beyond the scope of this paper; however, we briefly state several facts. Assuming the availability of constant time push and pop operations, the runtime is linear in the number of the nodes in the parse tree that is generated by the parser/generator (intuitively, this is true since although an individual \textsc{reduce} operation may require applying a number of pops that is linear in the number of input symbols, the total number of pop operations across an entire parse/generation run will also be linear). Since there is no way to bound the number of output nodes in a parse tree as a function of the number of input words, stating the runtime complexity of the parsing algorithm as a function of the input size requires further assumptions. Assuming our fixed constraint on maximum depth, it is linear.

\subsection{Comparison to Other Models}
Our generation algorithm algorithm differs from previous stack-based parsing/generation algorithms in two ways. First, it constructs rooted tree structures top down (rather than bottom up), and second, the transition operators are capable of directly generating arbitrary tree structures rather than, e.g., assuming binarized trees, as is the case in much prior work that has used transition-based algorithms to produce phrase-structure trees~\citep{sagae:2005,zhang:2011,zhu:2013}.
% Although, left-corner based algorithms likewise can generate arbitrary tree structures in a bottom-up order~\citep{henderson:2004}.

\section{Generative Model}\label{sec:model}
RNNGs use the generator transition set just presented to define a joint distribution on syntax trees ($\boldsymbol{y}$) and words ($\boldsymbol{x}$). This distribution is defined as a sequence model over generator transitions that is parameterized using a continuous space embedding of the algorithm state at each time step ($\mathbf{u}_t$); i.e.,
\begin{align*}
p(\boldsymbol{x},\boldsymbol{y}) &= \prod_{t=1}^{|\boldsymbol{a}(\boldsymbol{x},\boldsymbol{y})|} p(a_t \mid \boldsymbol{a}_{<t}) \\
&= \prod_{t=1}^{|\boldsymbol{a}(\boldsymbol{x},\boldsymbol{y})|}\frac{\exp \mathbf{r}^{\top}_{a_t} \mathbf{u}_t + b_{a_t}}{\sum_{a' \in \mathcal{A}_{G}(T_t,S_t,n_t)} \exp \mathbf{r}^{\top}_{a'} \mathbf{u}_t + b_{a'}},
\end{align*}
and where action-specific embeddings $\mathbf{r}_{a}$ and bias vector $\mathbf{b}$ are parameters in $\Theta$.

\begin{figure*}
\centering
\vspace{-.5cm}\includegraphics[scale=.6]{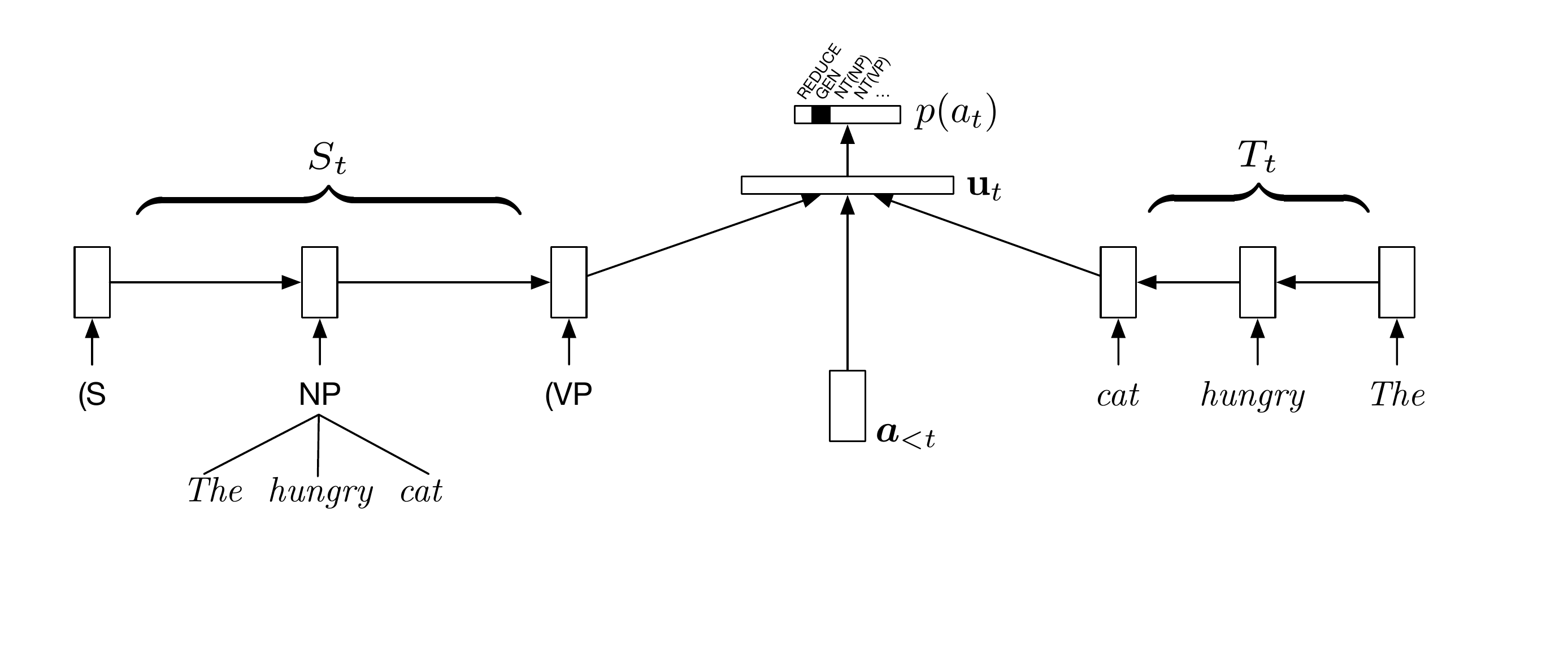}
\vspace{-2cm}\caption{Neural architecture for defining a distribution over $a_t$ given representations of the stack ($S_t$), output buffer ($T_t$) and history of actions ($\boldsymbol{a}_{<t}$). Details of the composition architecture of the NP, the action history LSTM, and the other elements of the stack are not shown. This architecture corresponds to the generator state at line 7 of Figure~\ref{fig:gen}.}
\label{fig:genstate}
\end{figure*}

The representation of the algorithm state at time $t$, $\mathbf{u}_t$, is computed by combining the representation of the generator's three data structures: the output buffer ($T_t$), represented by an embedding $\mathbf{o}_t$, the stack ($S_t$), represented by an embedding $\mathbf{s}_t$, and the history of actions ($\boldsymbol{a}_{<t}$) taken by the generator, represented by an embedding $\mathbf{h}_t$,
\begin{align*}
\mathbf{u}_t = \tanh\left( \mathbf{W} [\mathbf{o}_t ; \mathbf{s}_t ; \mathbf{h}_t ] + \mathbf{c} \right),
\end{align*}
where $\mathbf{W}$ and $\mathbf{c}$ are parameters. Refer to Figure~\ref{fig:genstate} for an illustration of the architecture.

The output buffer, stack, and history are sequences that grow unboundedly, and to obtain representations of them we use recurrent neural networks to ``encode'' their contents~\citep{cho:2014}. Since the output buffer and history of actions are only appended to and only contain symbols from a finite alphabet, it is straightforward to apply a standard RNN encoding architecture. The stack ($S$) is more complicated for two reasons. First, the elements of the stack are more complicated objects than symbols from a discrete alphabet: open nonterminals, terminals, and full trees, are all present on the stack. Second, it is manipulated using both push and pop operations. To efficiently obtain representations of $S$ under push and pop operations, we use stack LSTMs~\citep{dyer:2015}. To represent complex parse trees, we define a new syntactic composition function that recursively defines representations of trees.

\subsection{Syntactic Composition Function}
\begin{figure}[h]
\hspace{-.5cm}\includegraphics[scale=.5]{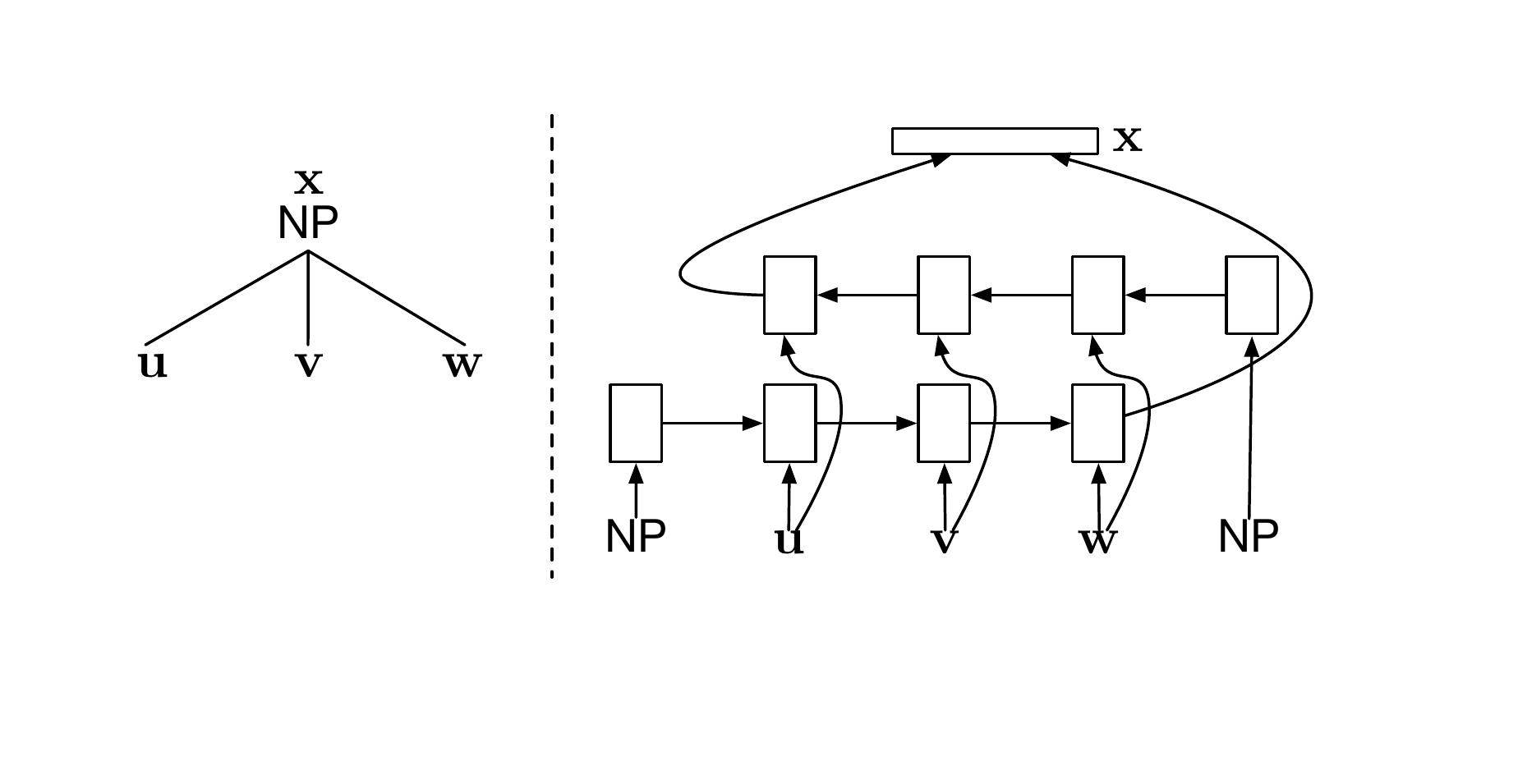}
\vspace{-2cm}
\caption{Syntactic composition function based on bidirectional LSTMs that is executed during a \textsc{reduce} operation; the network on the right models the structure on the left.}
\label{fig:composition}
\end{figure}

When a \textsc{reduce} operation is executed, the parser pops a sequence of completed subtrees and/or tokens (together with their vector embeddings) from the stack and makes them children of the most recent open nonterminal on the stack, ``completing'' the constituent. To compute an embedding of this new subtree, we use a composition function based on bidirectional LSTMs, which is illustrated in Fig.~\ref{fig:composition}.

The first vector read by the LSTM in both the forward and reverse directions is an embedding of the label on the constituent being constructed (in the figure, NP). This is followed by the embeddings of the child subtrees (or tokens) in forward or reverse order. Intuitively, this order serves to ``notify'' each LSTM what sort of head it should be looking for as it processes the child node embeddings. The final state of the forward and reverse LSTMs are concatenated, passed through an affine transformation and a $\tanh$ nonlinearity to become the subtree embedding.\footnote{We found the many previously proposed syntactic composition functions inadequate for our purposes. First, we must contend with an unbounded number of children, and many previously proposed functions are limited to binary branching nodes~\citep{socher:2013,dyer:2015}. Second, those that could deal with $n$-ary nodes made poor use of nonterminal information~\citep{tai:2015}, which is crucial for our task.} Because each of the child node embeddings ($\mathbf{u}$, $\mathbf{v}$, $\mathbf{w}$ in Fig.~\ref{fig:composition}) is computed similarly (if it corresponds to an internal node), this composition function is a kind of recursive neural network.

\subsection{Word Generation}
To reduce the size of $\mathcal{A}_G(S,T,n)$, word generation is broken into two parts. First, the decision to generate is made (by predicting $\textsc{gen}$ as an action), and then choosing the word, conditional on the current parser state. To further reduce the computational complexity of modeling the generation of a word, we use a class-factored softmax \citep{baltescu:2015,goodman:2001}. By using $\sqrt{|\Sigma|}$ classes for a vocabulary of size $|\Sigma|$, this prediction step runs in time $O(\sqrt{|\Sigma|})$ rather than the $O(|\Sigma|)$ of the full-vocabulary softmax. To obtain clusters, we use the greedy agglomerative clustering algorithm of~\citet{brown:1992}.

\subsection{Training}
The parameters in the model are learned to maximize the likelihood of a corpus of trees.

\subsection{Discriminative Parsing Model}\label{sec:disc}
A discriminative parsing model can be obtained by replacing the embedding of $T_t$ at each time step with an embedding of the input buffer $B_t$. To train this model, the conditional likelihood of each sequence of actions given the input string is maximized.\footnote{For the discriminative parser, the POS tags are processed similarly as in \citep{dyer:2015}; they are predicted for English with the Stanford Tagger \citep{Toutanova:2003:FPT:1073445.1073478} and Chinese with Marmot \citep{marmot}.}

\section{Inference via Importance Sampling}
\label{sec:inference}
Our generative model $p(\boldsymbol{x},\boldsymbol{y})$ defines a joint distribution on trees ($\boldsymbol{y}$) and sequences of words ($\boldsymbol{x}$). To evaluate this as a language model, it is necessary to compute the marginal probability $p(\boldsymbol{x}) = \sum_{\boldsymbol{y}' \in \mathcal{Y}(\boldsymbol{x})}p(\boldsymbol{x},\boldsymbol{y}')$. And, to evaluate the model as a parser, we need to be able to find the MAP parse tree, i.e., the tree $\boldsymbol{y} \in \mathcal{Y}(\boldsymbol{x})$ that maximizes $p(\boldsymbol{x},\boldsymbol{y})$. However, because of the unbounded dependencies across the sequence of parsing actions in our model, exactly solving either of these inference problems is intractable. To obtain estimates of these, we use a variant of importance sampling \citep{doucet:2011}.
\ignore{\footnote{\citet{buys:2015a} proposed using particle filtering to estimate the marginal probabilities and obtain an estimate of the MAP parse tree. While their approach could, with adaptations to deal with the $\textsc{nt}(\mathrm{X})$ operations, be used with our model as well, we take a slightly different view of the problem and use a discriminatively trained parser to assist in computing the marginal likelihood of sentences under our model. Our algorithm has two advantages over the particle filtering approach. First, since our proposal distribution conditions on all of $\boldsymbol{x}$ when predicting actions---whereas the Buys and Blunsom particle filter only considers the words that have been generated---this can be understood as a variance reduction strategy, and variance is a known limitation of particle filters.}}

Our importance sampling algorithm uses a conditional proposal distribution $q(\boldsymbol{y} \mid \boldsymbol{x})$ with the following properties: (i)~$p(\boldsymbol{x},\boldsymbol{y})>0 \implies q(\boldsymbol{y} \mid \boldsymbol{x})>0$; (ii)~samples $\boldsymbol{y} \sim q(\boldsymbol{y} \mid \boldsymbol{x})$ can be obtained efficiently; and (iii)~the conditional probabilities $q(\boldsymbol{y} \mid \boldsymbol{x})$ of these samples are known. While many such distributions are available, the discriminatively trained variant of our parser (\S\ref{sec:disc}) fulfills these requirements: sequences of actions can be sampled using a simple ancestral sampling approach, and, since parse trees and action sequences exist in a one-to-one relationship, the product of the action probabilities is the conditional probability of the parse tree under $q$. We therefore use our discriminative parser as our proposal distribution.

%An example of a reasonable $q$ might be, e.g., a simple PCFG which admits perfect sampling via the well-known tree-structured generalization of the forward-filtering/backward-sampling algorithm \citep[\emph{inter alia}]{goodman:1998,johnson:2007}.

Importance sampling uses \textbf{importance weights}, which we define as $w(\boldsymbol{x},\boldsymbol{y}) = p(\boldsymbol{x},\boldsymbol{y})/q(\boldsymbol{y} \mid \boldsymbol{x})$, to compute this estimate. Under this definition, we can derive the estimator as follows:
\begin{align*}
p(\boldsymbol{x}) &= \sum_{\boldsymbol{y} \in \mathcal{Y}(\boldsymbol{x})} p(\boldsymbol{x},\boldsymbol{y})  = \sum_{\boldsymbol{y} \in \mathcal{Y}(\boldsymbol{x})} q(\boldsymbol{y} \mid\boldsymbol{x})w(\boldsymbol{x},\boldsymbol{y}) \\
 &= \mathbb{E}_{q(\boldsymbol{y} \mid \boldsymbol{x})} w(\boldsymbol{x},\boldsymbol{y}).
\end{align*}
We now replace this expectation with its Monte Carlo estimate as follows, using $N$ samples from $q$:
\begin{align*}
\boldsymbol{y}^{(i)} \sim q(\boldsymbol{y} \mid \boldsymbol{x}) \quad &\textrm{for }i \in \{ 1, 2, \ldots, N\} \\
\mathbb{E}_{q(\boldsymbol{y} \mid \boldsymbol{x})} w(\boldsymbol{x},\boldsymbol{y}) &
%\approx \hat{\mathbb{E}}^{\textrm{MC}}_{q(\boldsymbol{y} \mid \boldsymbol{x})} w(\boldsymbol{x},\boldsymbol{y}) \\ &
\stackrel{\textrm{MC}}{\approx} \frac{1}{N} \sum_{i=1}^N w(\boldsymbol{x},\boldsymbol{y}^{(i)}) \\
\end{align*}
To obtain an estimate of the MAP tree $\hat{\boldsymbol{y}}$, we choose the sampled tree with the highest probability under the joint model $p(\boldsymbol{x},\boldsymbol{y})$.

\section{Experiments} \label{sec:exp}

We present results of our two models both on parsing (discriminative and generative) and as a language model (generative only) in English and Chinese.

\paragraph{Data.} For English, \S 2--21 of the Penn Treebank are used as training corpus for both, with {\S}24 held out as validation, and {\S23} used for evaluation. Singleton words in the training corpus with unknown word classes using the the Berkeley parser's mapping rules.\footnote{\url{http://github.com/slavpetrov/berkeleyparser}} Orthographic case distinctions are preserved, and numbers (beyond singletons) are not normalized. For Chinese, we use the Penn Chinese Treebank Version~5.1 (CTB) \citep{Xue:2005:PCT:1064781.1064785}.\footnote{\S 001--270 and 440--1151 for training, \S 301--325 development data, and \S 271--300 for evaluation.} For the Chinese experiments, we use a single unknown word class. Corpus statistics are given in Table~\ref{tab:ptb}.\footnote{This preprocessing scheme is more similar to what is standard in parsing than what is standard in language modeling. However, since our model is both a parser and a language model, we opted for the parser normalization.} %We pre-train the word embedding of \citep{Ling:2015} for both English (64-dimensional) and Chinese (50-dimensional Gigaword) and compose them during training with the learned word embedding (in addition to POS-tag embedding for the discriminative model) in a similar manner as \citep{dyer:2015}.

\begin{table}[h]
\caption{Corpus statistics.}
\begin{center}
\begin{scriptsize}
\begin{tabular}{l|r|r|r|r}
 & PTB-train & PTB-test & CTB-train & CTB-test \\
\hline
Sequences & 39,831 &2,416 & 50,734 & 348 \\
Tokens & 950,012 & 56,684 &1,184,532 & 8,008\\
Types & 23,815 & 6,823 & 31,358 &1,637 \\
UNK-Types & 49 & 42 & 1 & 1 \\
\end{tabular}
\end{scriptsize}
\end{center}
\label{tab:ptb}
\end{table}%

\paragraph{Model and training parameters.} For the discriminative model, we used hidden dimensions of 128 and 2-layer LSTMs (larger numbers of dimensions reduced validation set performance). For the generative model, we used 256 dimensions and 2-layer LSTMs. For both models, we tuned the dropout rate to maximize validation set likelihood, obtaining optimal rates of 0.2 (discriminative) and 0.3 (generative). For the sequential LSTM baseline for the language model, we also found an optimal dropout rate of 0.3. For training we used stochastic gradient descent with a learning rate of 0.1. All parameters were initialized according to recommendations given by \citet{glorot:2010}.

\paragraph{English parsing results.} Table~\ref{tab:parsing} (last two rows) gives the performance of our parser on Section 23, as well as the performance of several representative models. For the discriminative model, we used a greedy decoding rule as opposed to beam search in some shift-reduce baselines. For the generative model, we obtained 100 independent samples from a flattened distribution of the discriminative parser (by exponentiating each probability by $\alpha=0.8$ and renormalizing) and reranked them according to the generative model.\footnote{The value $\alpha=0.8$ was chosen based on the diversity of the samples generated on the development set.}

\begin{table}[h]
\caption{Parsing results on PTB \S 23 (D=discriminative, G=generative,  S=semisupervised). $^{\star}$ indicates the (Vinyals et al., 2015) result with trained only on the WSJ corpus without ensembling.}\begin{center}
\begin{tabular}{l|c|c}
\textbf{Model} & \textbf{type} & \textbf{F}$_\mathbf{1}$ \\
\hline
% \citet{sagae:2005} & D & 87.6 \\
\citet{vinyals:2015}$^{\star}$ -- WSJ only & D & 88.3 \\
\citet{henderson:2004} & D & 89.4 \\
\citet{socher:2013parsing} & D & 90.4 \\
\citet{zhu:2013} & D & 90.4 \\
%\citet{carreras-collins-koo:2008:CONLL} & D & 91.1 \\
%\citet{durrett:2015} & D & 91.1 \\
\hline
\citet{petrov:2007} & G & 90.1 \\
\citet{bod:2003} & G & 90.7 \\
\citet{shindo:2012} -- single & G & 91.1 \\
\citet{shindo:2012} -- ensemble & G & 92.4 \\
\hline
\citet{zhu:2013} & S & 91.3 \\
\citet{mcclosky:2006} & S & 92.1 \\
\citet{vinyals:2015} -- single & S & 92.1 \\
%\citet{vinyals:2015} -- ensemble & S & 92.8 \\
\hline
\hline
Discriminative, $q(\boldsymbol{y} \mid \boldsymbol{x})$ & D & 89.8  \\
Generative, $\hat{p}(\boldsymbol{y} \mid \boldsymbol{x})$ & G & 92.4 \\
\end{tabular}
\end{center}
\label{tab:parsing}
\end{table}

\paragraph{Chinese parsing results.} Chinese parsing results were obtained with the same methodology as in English and show the same pattern (Table~\ref{tab:ctbparse}).

\begin{table}[h]
\begin{center}
\caption{Parsing results on CTB 5.1.}
\begin{tabular}{l|c|c}
\textbf{Model} & \textbf{type} & \textbf{F}$_\mathbf{1}$ \\
\hline
%\citet{charniak:2005} & D & 82.3 \\
\citet{zhu:2013} & D & 82.6 \\
%\citet{wang-xue:2014:P14-1} & D & 82.7\\
\citet{wang-mi-xue:2015:ACL-IJCNLP} & D & 83.2\\
\citet{Huang:2009:SPG:1699571.1699621} & D & 84.2 \\
\hline
\citet{charniak:2000} & G & 80.8 \\
\citet{bikel:2004} & G & 80.6 \\
\citet{petrov:2007} & G & 83.3 \\
\hline
%\citet{Huang:2009:SPG:1699571.1699621} & S & 85.2 \\
%\citet{Wang11} & S & 86.6\\
\citet{zhu:2013} & S & 85.6 \\
\citet{wang-xue:2014:P14-1} & S & 86.3\\
\citet{wang-mi-xue:2015:ACL-IJCNLP} & S & 86.6\\
\hline
\hline
Discriminative, $q(\boldsymbol{y} \mid \boldsymbol{x})$ & D & 80.7 \\
Generative, $\hat{p}(\boldsymbol{y} \mid \boldsymbol{x})$ & G & 82.7 \\
\end{tabular}
\end{center}
\label{tab:ctbparse}
\end{table}

%These results are quite surprising. First, although the discriminative model has access to more information than the generative model, the generative model substantially outperforms the discriminative model. Indeed, the performance of the discriminative model is remarkably bad, while the performance of the generative model is exceeds that of any prior published generative parser. These results are perhaps more surprising in that random samples from the discriminative model are ``reranked'' by the generative model (\S\ref{sec:inference}).

\paragraph{Language model results.} We report held-out per-word perplexities of three language models, both sequential and syntactic.
%To ensure that these quantities are exactly comparable, we use the same vocabulary, and consider the probability of sequential models to include the end-of-sentence marker.
%Since the recursive rewriting process of the syntactic model terminates with probability 1 when a final parsing state is reached, this symbol is only implicitly considered in the syntactic model.
Log probabilities are normalized by the number of words (excluding the stop symbol), inverted, and exponentiated to yield the perplexity. Results are summarized in Table~\ref{tab:ppl}. %Random samples of text from the sequential LSTM model and the RNNG-based model are shown in Figure~\ref{fig:samples}.

\ignore{\begin{table}[h]
\caption{Language model perplexity results.}
\begin{center}
\begin{tabular}{l|r|r}
Model & dev ppl & test ppl \\
\hline
IKneser--Ney 5-gram & 209.7 & 169.3 \\
LSTM & 162.1 & 131.7   \\
LSTM + Dropout & 139.5 & 113.4   \\
LSTM Syntax &  & 125.5 \\
LSTM Syntax + Dropout &  & 102.4
\end{tabular}
\end{center}
\label{tab:ppl}
\end{table}}

\begin{table}[h]
\caption{Language model perplexity results.}
\begin{center}
\begin{tabular}{l|c|c}
\textbf{Model} & \textbf{test ppl} (PTB) & \textbf{test ppl} (CTB) \\
\hline
IKN 5-gram & 169.3 & 255.2\\
LSTM LM & 113.4  & 207.3 \\
RNNG &  102.4 & 171.9
\end{tabular}
\end{center}
\label{tab:ppl}
\end{table}%

\section{Discussion}
It is clear from our experiments that the proposed generative model is quite effective both as a parser and as a language model.  This is the result of (i) relaxing conventional independence assumptions (e.g., context-freeness) and (ii) inferring continuous representations of symbols alongside non-linear models of their syntactic relationships. The most significant question that remains is why the discriminative model---which has more information available to it than the generative model---performs worse than the generative model. This pattern has been observed before in neural parsing by \citet{henderson:2004}, who hypothesized that larger, unstructured conditioning contexts are harder to learn from, and provide opportunities to overfit. Our discriminative model conditions on the entire history, stack, and buffer, while our generative model only accesses the history and stack.
The fully discriminative model of \citet{vinyals:2015} was able to obtain results similar to those of our generative model (albeit using much larger training sets obtained through semisupervision) but similar results to those of our discriminative parser using the same data.  In light of their results, we believe Henderson's hypothesis is correct, and that generative models should be considered as a more statistically efficient method for learning neural networks from small data.

\section{Related Work}
Our language model combines work from two modeling traditions: (i)~recurrent neural network language models and (ii)~syntactic language modeling. Recurrent neural network language models use RNNs to compute representations of an unbounded history of words in a left-to-right language model~\citep{zaremba:2015b,mikolov:2010,elman:1990}. Syntactic language models jointly generate a syntactic structure and a sequence of words \citep{baker-79,jelinek-91}. There is an extensive literature here, but one strand of work has emphasized a bottom-up generation of the tree, using variants of shift-reduce parser actions to define the probability space \citep{chelba:2000,emami:2005}. The neural-network--based model of \citet{henderson:2004} is particularly similar to ours in using an unbounded history in a neural network architecture to parameterize generative parsing based on a left-corner model. Dependency-only language models have also been explored \citep{titov:2007b,buys:2015a,buys:2015b}.
Modeling generation top-down as a rooted branching process that recursively rewrites nonterminals has been explored by \citet{charniak:2000} and \citet{roark:2001}. Of particular note is the work of \citet{charniak:2010}, which uses random forests and hand-engineered features over the entire syntactic derivation history to make decisions over the next action to take.

 %Although most of this work has used hand-engineered features, \citet{black:1993} used decision trees to learn to condition on arbitrary aspects of a top-down generative history.

The neural networks we use to model sentences are structured according to the syntax of the sentence being generated. Syntactically structured neural architectures have been explored in a number of applications, including discriminative parsing~\citep{socher:2013parsing,kiperwasser:2016}, sentiment analysis~\citep{tai:2015,socher:2013}, and sentence representation~\citep{socher:2011,bowman:2016}. However, these models have been, without exception, discriminative; this is the first work to use syntactically structured  neural models to generate language.  Earlier work has demonstrated that sequential RNNs have the capacity to recognize context-free (and beyond) languages~\citep{sun-98,siegelmann:1995}. In contrast, our work may be understood as a way of incorporating a context-free inductive bias into the model structure.

\section{Outlook}
RNNGs can be combined with a particle filter inference scheme (rather than the importance sampling method based on a discriminative parser, \S\ref{sec:inference}) to produce a left-to-right marginalization algorithm that runs in expected linear time. Thus, they could be used in applications that require language models.

A second possibility is to replace the sequential generation architectures found in many neural network transduction problems that produce sentences conditioned on some input. Previous work in machine translation has showed that conditional syntactic models can function quite well without the computationally expensive marginalization process at decoding time~\citep{galley:2006,gimpel-14}.

A third consideration regarding how RNNGs, human sentence processing takes place in a left-to-right, incremental order. While an RNNG is not a processing model (it is a grammar), the fact that it is left-to-right opens up several possibilities for developing new sentence processing models based on an explicit grammars, similar to the processing model of \citet{charniak:2010}.

Finally, although we considered only the supervised learning scenario, RNNGs are joint models that could be trained without trees, for example, using expectation maximization.

\section{Conclusion}
We introduced recurrent neural network grammars, a probabilistic model of phrase-structure trees that can be trained generatively and used as a language model or a parser, and a corresponding discriminative model that can be used as a parser.  Apart from out-of-vocabulary preprocessing, the approach requires no feature design or transformations to treebank data.  The generative model outperforms every previously published parser built on a single supervised generative model in English, and a bit behind the best-reported generative model in Chinese. As language models, RNNGs outperform the best single-sentence language models.

\section*{Acknowledgments}
We thank Brendan O'Connor, Swabha Swayamdipta, and Brian Roark for feedback on drafts of this paper, and Jan Buys, Phil Blunsom, and Yue Zhang for help with data preparation. This work was sponsored in part by the Defense Advanced Research Projects Agency (DARPA)
Information Innovation Office (I2O) under the Low Resource Languages for Emergent Incidents (LORELEI) program issued by DARPA/I2O under Contract No.~HR0011-15-C-0114;
it was also supported in part by Contract No.~W911NF-15-1-0543 with the DARPA and the Army Research Office (ARO). Approved for public release, distribution unlimited. The views expressed are those of the authors and do not reflect the official policy or position of the Department of Defense or the U.S.~Government.
Miguel Ballesteros was supported by the European Commission under the contract numbers FP7-ICT-610411 (project MULTISENSOR) and H2020-RIA-645012 (project KRISTINA).

\setlength{\bibsep}{2pt}
{\fontsize{10}{12.25}\selectfont
\bibliography{synlm}}

%\bibliography{biblio}

%
% File emnlp2016.tex
%

%\documentclass[11pt,letterpaper]{article}
%\usepackage{emnlp2016}
%\usepackage{times}
%\usepackage{latexsym}
%\usepackage{graphicx}
%\usepackage{kantlipsum}
%\usepackage{color}
%\usepackage{amsmath,array}

% Uncomment this line for the final submission:
%\emnlpfinalcopy

%\newcommand{\mbcomment}[1]{\textcolor{red}{\textbf{[#1 --\textsc{mb}]}}}
%\newcommand{\akcomment}[1]{\textcolor{blue}{\textbf{[#1 --\textsc{mb}]}}}

%  Enter the EMNLP Paper ID here:
%\def\emnlppaperid{***}

% To expand the titlebox for more authors, uncomment
% below and set accordingly.
% \addtolength\titlebox{.5in}    

%\newcommand\BibTeX{B{\sc ib}\TeX}

\newpage

\author{\textbf{Chris Dyer}$^{\spadesuit\clubsuit}$ ~ \textbf{Adhiguna Kuncoro}$^{\spadesuit}$ ~ \textbf{Miguel Ballesteros}$^{\diamondsuit}$   ~ \textbf{Noah A. Smith}$^{\heartsuit}$ \\
$^{\spadesuit}$School of Computer Science, Carnegie Mellon University, Pittsburgh, PA, USA \\
$^{\diamondsuit}$NLP Group, Pompeu Fabra University, Barcelona, Spain \\
$^{\clubsuit}$Google DeepMind, London, UK\\
$^{\heartsuit}$Computer Science \& Engineering, University of Washington, Seattle, WA, USA\\
{\small \tt \{cdyer,akuncoro\}@cs.cmu.edu}\\ {\small \tt miguel.ballesteros@upf.edu, nasmith@cs.washington.edu}
}
\date{}

%\begin{document}
%\maketitle
\twocolumn[\subsubsection*{\centering Corrigendum to Recurrent Neural Network Grammars}\vspace{15 mm}]

 \begin{abstract}
Due to an implentation bug in the RNNG's recursive composition function, the results reported in Dyer et al. (2016) did not correspond to the model as it was presented. This corrigendum describes the buggy implementation and reports results with a corrected implementation. After correction, on the PTB \S23 and CTB 5.1 test sets, respectively, the generative model achieves language modeling perplexities of 105.2 and 148.5, and phrase-structure parsing F1 of 93.3 and 86.9, a new state of the art in phrase-structure parsing for both languages.
%We outline a correction to the Recurrent Neural Network Grammars (RNNG) regarding an implementation mistake on the original publication. The fixed model achieved a language modeling perplexity of 105.2 and constituency parsing F1 of 93.3 on the Penn Treebank Section 23 test set, a new state of the art in constituency parsing.
\end{abstract}

%\section{Introduction}
%In \newcite{rnng}, we introduced recurrent neural network grammars (RNNG), a probabilistic model of natural language based on phrase-structure trees. The RNNG is a transition-based generative model of language that \emph{jointly} models strings and trees using recurrent neural networks, breaking down each sentence into a sequence of \emph{actions}.

%The overall architecture of the RNNG is summarized in Figure \ref{fig:rnng_overall}. The RNNG consists of a \emph{stack} of partially constructed constituency trees, an \emph{output buffer} of all the terminal symbols generated so far, and a list of \emph{past actions} that provides information on all past actions that led the model to the current state. Each component is encoded by a separate LSTM network, with a composition (i.e. reduce operation) in order to get the embedding of a completed constituent on the stack, as described in \S\ref{sec:composition}. Decoding with the RNNG model used an importance sampling mechanism by sampling trees from a discriminative model of the RNNG. We refer interested readers to the original publication for more details.

%\begin{figure*}[h]
%\centering
%\includegraphics[scale=0.45]{rnng_overall.eps}
%\caption{Architecture of the RNNG)}
%\label{fig:rnng_overall}
%\end{figure*}

\section*{RNNG Composition Function and Implementation Error}\label{sec:composition}
\begin{figure*}
\centering
\begin{minipage}{.46\textwidth}
  \centering
  \includegraphics[width=.9\linewidth]{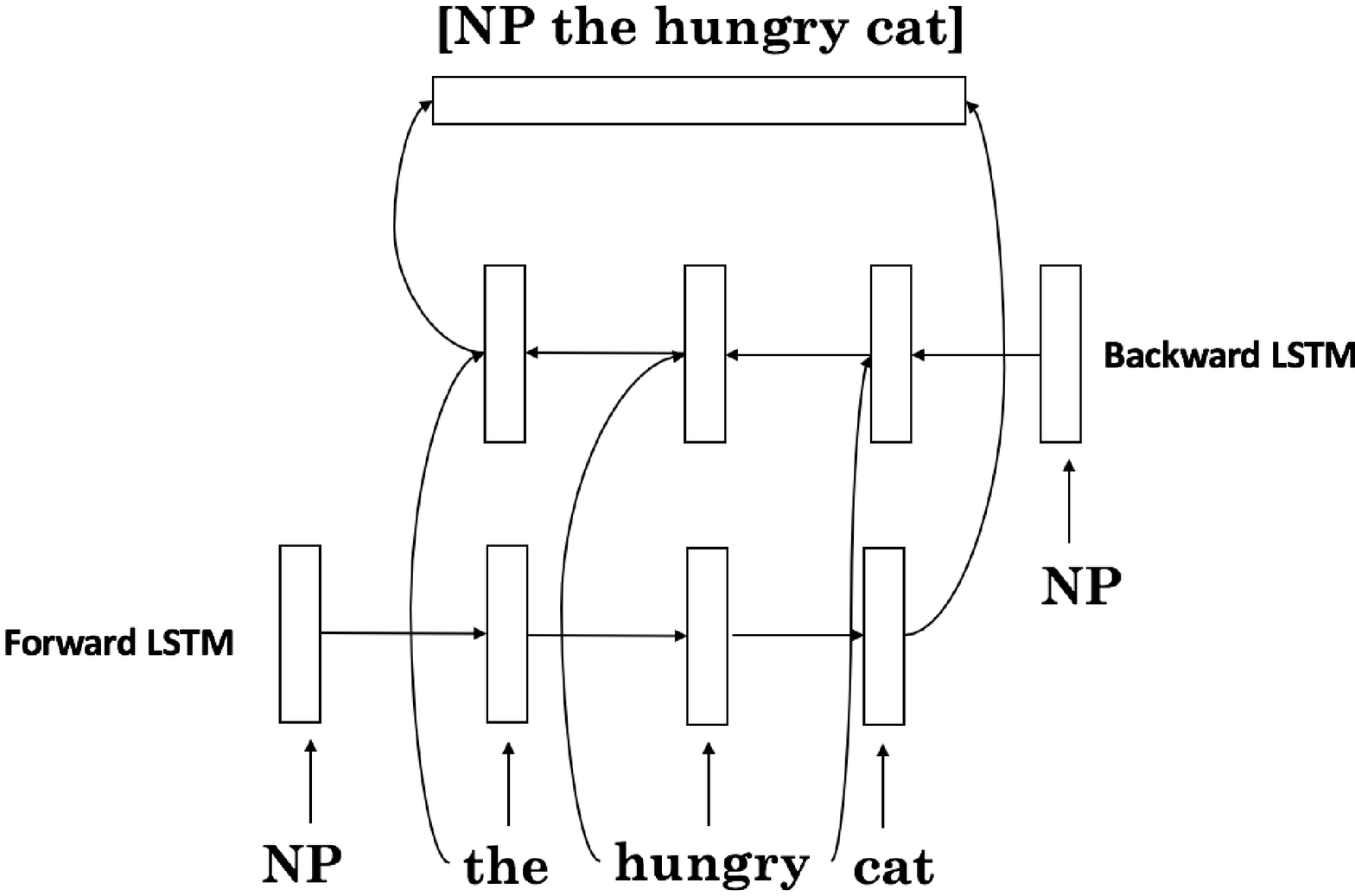}
  \caption{Correct RNNG composition function for the constituent (NP \emph{the hungry cat}).}
  \label{fig:composition_ideal}
\end{minipage}%
\begin{minipage}{.04\textwidth}${}$\end{minipage}
\begin{minipage}{.46\textwidth}
  \centering
  \includegraphics[width=.9\linewidth]{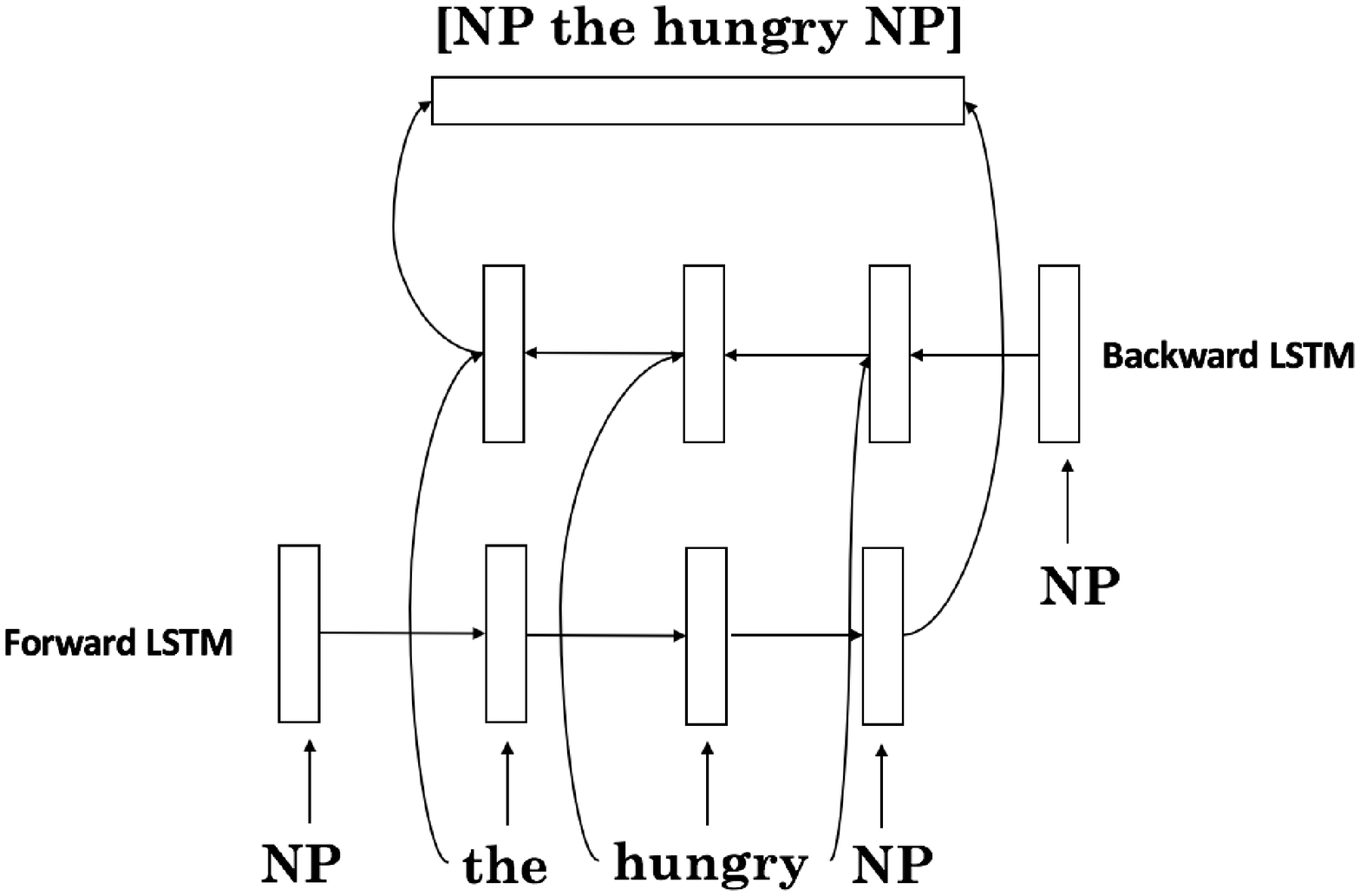}
  \caption{Buggy implementation of the RNNG composition function for the constituent (NP \emph{the hungry cat}).  Note that the right-most child, \emph{cat}, has been replaced by a second NP.}
  \label{fig:composition_broken}
\end{minipage}
\end{figure*}

The composition function reduces a completed constituent into a single vector representation using a bidirectional LSTM (Figure \ref{fig:composition_ideal}) over embeddings of the constituent's children as well as an embedding of the resulting nonterminal symbol type. The implementation error (Figure \ref{fig:composition_broken}) composed the constituent (NP \emph{the hungry cat}) by reading the sequence ``NP \emph{the hungry} NP'', that is, it discarded the rightmost child of every constituent and replaced it with a second copy of the constituent's nonterminal symbol. This error occurs for every constituent and means crucial information is not properly propagated upwards in the tree. %Both forward and backward LSTMs first read the non-terminal symbol, followed by the constituent children in each respective direction. %Intuitively, providing the non-terminal symbol as the first input to both LSTMs serve to notify the model on what sort of children they should attend to (e.g. for noun phrases this might be the rightmost noun since it is typically considered the lexical head). 
%The composition function for a noun phrase ``the hungry cat'' is illustrated in Figure \ref{fig:composition_ideal}.

%\mbcomment{Please, draw the figures as in the RNNG pape, using latex or graffle... in png or jpg they normally do not look very nice.}

%\section{Implementation Mistake on Composition Function}
%In the case of the sample noun phrase ``NP the hungry cat'', the terminal ``cat'', widely deemed to be the lexical head and most important element of the noun phrase, is not being composed as part of the noun phrase. Note that this issue is potentially mitigated by the existence of the output buffer that records all generated terminals so far, which explains how the RNNG achieved state of the art accuracy despite this implementation error. We fixed this issue by modifying the composition function to reflect the original design.

\section*{Results after Correction}
The implementation error affected both the generative and discriminative RNNGs.\footnote{The discriminative model can only be used for parsing and not for language modeling, since it only models $p(\boldsymbol{y} \mid \boldsymbol{x})$.} We summarize corrected English phrase-structure PTB \S23 parsing result in Table \ref{tab:parsing_correct}, Chinese (CTB 5.1 \S271--300) in Table \ref{tab:parsing_chinese_correct} (achieving the the best reported result on both datasets), and English and Chinese language modeling perplexities in Table \ref{tab:lm_correct}. The considerable improvement in parsing accuracy indicates that properly composing the constituent and propagating information upwards is crucial. Despite slightly higher language modeling perplexity on PTB \S23, the fixed RNNG still outperforms a highly optimized sequential LSTM baseline.

\begin{table}[h]
\begin{center}
\begin{tabular}{l|c|c}
\textbf{Model} & \textbf{type} & \textbf{\em F}$_\mathbf{1}$ \\
\hline
% \citet{sagae:2005} & D & 87.6 \\
\cite{vinyals:2015}$^{\star}$ -- WSJ only & D & 88.3 \\
\cite{henderson:2004} & D & 89.4 \\
\cite{socher:2013parsing} & D & 90.4 \\
\cite{zhu:2013} & D & 90.4 \\
%\citet{carreras-collins-koo:2008:CONLL} & D & 91.1 \\
%\citet{durrett:2015} & D & 91.1 \\
\hline
\cite{petrov:2007} & G & 90.1 \\
\cite{bod:2003} & G & 90.7 \\
\cite{shindo:2012} -- single & G & 91.1 \\
\cite{shindo:2012} -- ensemble & G & 92.4 \\
\hline
\cite{zhu:2013} & S & 91.3 \\
\cite{mcclosky:2006} & S & 92.1 \\
\cite{vinyals:2015} & S & 92.1 \\
%\cite{vinyals:2015} -- ensemble & S & 92.8 \\
\hline
\hline
Discriminative, $q(\boldsymbol{y} \mid \boldsymbol{x})^{\dagger}$ -- buggy & D & 89.8  \\
Generative, $\hat{p}(\boldsymbol{y} \mid \boldsymbol{x})^{\dagger}$ -- buggy & G & 92.4 \\
\hline
\hline
Discriminative, $q(\boldsymbol{y} \mid \boldsymbol{x})$ -- correct & D & 91.7  \\
Generative, $\hat{p}(\boldsymbol{y} \mid \boldsymbol{x})$ -- correct & G & \textbf{93.3} \\
\end{tabular}
\end{center}
\caption{Parsing results with fixed composition function on PTB \S 23 (D=discriminative, G=generative,  S=semisupervised). $^{\star}$ indicates the (Vinyals et al., 2015) model trained only on the WSJ corpus without ensembling. $^{\dagger}$ indicates RNNG models with the buggy composition function implementation.}
\label{tab:parsing_correct}
\end{table}

\begin{table}[h]
\begin{center}
\begin{tabular}{l|c|c}
\textbf{Model} & \textbf{type} & \textbf{F}$_\mathbf{1}$ \\
\hline
%\citet{charniak:2005} & D & 82.3 \\
\cite{zhu:2013} & D & 82.6 \\
%\citet{wang-xue:2014:P14-1} & D & 82.7\\
\cite{wang-mi-xue:2015:ACL-IJCNLP} & D & 83.2\\
\cite{Huang:2009:SPG:1699571.1699621} & D & 84.2 \\
\hline
\cite{charniak:2000} & G & 80.8 \\
\cite{bikel:2004} & G & 80.6 \\
\cite{petrov:2007} & G & 83.3 \\
\hline
%\citet{Huang:2009:SPG:1699571.1699621} & S & 85.2 \\
%\citet{Wang11} & S & 86.6\\
\cite{zhu:2013} & S & 85.6 \\
\cite{wang-xue:2014:P14-1} & S & 86.3\\
\cite{wang-mi-xue:2015:ACL-IJCNLP} & S & 86.6\\
\hline
\hline
Discriminative, $q(\boldsymbol{y} \mid \boldsymbol{x})^{\dagger}$ - buggy & D & 80.7 \\
Generative, $\hat{p}(\boldsymbol{y} \mid \boldsymbol{x})^{\dagger}$ - buggy & G & 82.7 \\
\hline
\hline
Discriminative, $q(\boldsymbol{y} \mid \boldsymbol{x})$ -- correct & D & 84.6  \\
Generative, $\hat{p}(\boldsymbol{y} \mid \boldsymbol{x})$ -- correct & G & \textbf{86.9}
\end{tabular}
\end{center}
\caption{Parsing results on CTB 5.1 including results with the buggy composition function implementation (indicated by $^{\dagger}$) and with the correct implementation.}
\label{tab:parsing_chinese_correct}
\end{table}

%We summarize language modeling accuracy using the generative RNNG on Table \ref{tab:lm}. The fixed composition function results in slightly higher language modeling perplexity, although it still outperforms a highly optimized sequential LSTM baseline that achieved a perplexity of 113.4. Note that we do not tune the fixed RNNG as exhaustively as the initial implementation. \mbcomment{I would say that these 3 points in perplexity is in the range of noise... }\akcomment{I think we cannot make this claim since we did not do a statistical test. The adjective ``slightly higher'' should be enough to emphasize that the difference is minimal}

\begin{table}[]
\begin{center}
\begin{tabular}{l|c|c}
\textbf{Model} & \textbf{test ppl} (PTB) & \textbf{test ppl} (CTB) \\
\hline
IKN 5-gram & 169.3 & 255.2 \\
LSTM LM & 113.4   & 207.3 \\
RNNG -- buggy$^{\dagger}$ &  \textbf{102.4} & 171.9 \\
RNNG -- correct &  105.2 & \textbf{148.5}\\ \hline
\end{tabular}
\end{center}
\caption{PTB and CTB language modeling results including results with the buggy composition function implementation (indicated by $^{\dagger}$) and with the correct implementation.}
\label{tab:lm_correct}
\end{table}

%\mbcomment{I would put the same table of results as in the RNNG paper, but with the update...}\akcomment{resolved}

%\section*{Acknowledgments}

%Do not number the acknowledgment section.

%\bibliography{emnlp2016}
%\bibliographystyle{emnlp2016}

%\end{document}

\end{document}